\documentclass[a4paper,fleqn]{cas-sc}

\usepackage[authoryear,longnamesfirst]{natbib}
\setcitestyle{numbers,square}
\usepackage{tabularx} 
\usepackage{multirow} 
\usepackage{subcaption} 
\graphicspath{{figs/}} 
\usepackage{amsmath}
\usepackage{algorithm}
\usepackage{algpseudocode}
\usepackage{amsfonts}
\def\tsc#1{\csdef{#1}{\textsc{\lowercase{#1}}\xspace}}
\tsc{WGM}
\tsc{QE}

\begin{document}
\let\WriteBookmarks\relax
\def\floatpagepagefraction{1}
\def\textpagefraction{.001}

\shorttitle{}    

\shortauthors{Wang J et al.}  

\title [mode = title]{Graph Similarity Computation via Interpretable Neural Node Alignment}  



%

\author[1]{Jingjing Wang}[orcid=0000-0002-6820-4777]
\ead{Wangj3573@163.com}

\ead[url]{}

\credit{Conceptualization, Methodology}
\affiliation[1]{organization={School of Computer Science, Hangzhou Dianzi University},
            addressline={1158 2nd Ave, Qiantang district}, 
            city={Hangzhou},
            postcode={310005}, 
            state={Zhejiang},
            country={China}}

\author[1]{Hongjie Zhu}
\ead{zhj_0221@163.com}
\credit{Visualization, Investigation}
\ead[url]{}

\author[2]{Haoran Xie}[orcid=0000-0003-0965-3617]
\credit{Validation}
\ead{hrxie@ln.edu.hk}
\ead[url]{https://xiehaoran.net/}

\author[3]{Fu Lee Wang}[orcid=0000-0002-3976-0053]
\credit{Writing-Reviewing and Editing}
\ead{pwang@hkmu.edu.hk}
\ead[url]{https://scholars.hkmu.edu.hk/en/persons/fu-lee-wang}

\author[1]{Xiaoliang Xu}
\cormark[1]
\ead{xxl@hdu.edu.cn}
\credit{Supervision}
\author[1]{Yuxiang Wang}
\ead{lsswyx@hdu.edu.cn}
\credit{Data curation}
\affiliation[2]{organization={School of Data Science, Lingnan University},
            addressline={8 Castle Peak Road}, 
            city={Tuen Mun},
            postcode={999097}, 
            state={New Territories},
            country={Hong Kong SAR}}
\affiliation[3]{organization={School of Science and Technology, Hong Kong Metropolitan University},
            addressline={30 Good Shepherd Street}, 
            city={Ho Man Tin},
            postcode={999097}, 
            state={Kowloon},
            country={Hong Kong SAR}}
\cortext[1]{Corresponding author}



\begin{abstract}
Graph similarity computation is an essential task in many real-world graph-related applications such as retrieving the similar drugs given a query chemical compound or finding the user's potential friends from the social network database. Graph Edit Distance (GED) and Maximum Common Subgraphs (MCS) are the two commonly used domain-agnostic metrics to evaluate graph similarity in practice, by quantifying the minimum cost under optimal alignment between the entities of two graphs, such as nodes, edges, and subgraphs. Unfortunately, computing the exact GED is known to be a NP-hard problem. To solve this limitation, neural network based models have been proposed to approximate the calculations of GED/MCS between the given pair of graphs. However, deep learning models are well-known 'black boxes', thus the typically characteristic one-to-one node/subgraph alignment process in the classical computations of GED and MCS cannot be seen. Such alignment information provides interpretability for graph editing operations and visualizes the common subgraph between two graphs, which is extremely useful for graph-based downstream tasks. Existing methods have paid attention to approximating the node/subgraph alignment (soft alignment), but the one-to-one node alignment (hard alignment) has not yet been solved. To fill this gap, in this paper we propose a novel interpretable neural node alignment model without relying on node alignment ground truth information. Firstly, the quadratic assignment problem in classical GED computation is relaxed to a linear alignment via embedding the features in the node embedding space. Secondly, a differentiable Gumbel-Sinkhorn module is proposed to unsupervised generate the optimal one-to-one node alignment matrix. Experimental results in real-world graph datasets demonstrate that our method outperforms the state-of-the-art methods in graph similarity computation and graph retrieval tasks, achieving up to 16\% reduction in the Mean Squared Error and up to 12\% improvement in the retrieval evaluation metrics, respectively. Furthermore, the experimental results demonstrate that interpretability and accuracy are not always opposites, in fact, enhancing interpretability can benefit accuracy.
The source code is available at \url{https://github.com/sasr11/GNA}.
\end{abstract}




\maketitle

\section{Introduction}\label{sec-1}
Real-world data are usually nonlinear structures, such as social networks in sociology, protein-protein interactions in natural sciences, and user behavior histories in recommendation systems. The complex relationships between nodes within a graph provide rich semantic information, making graphs a hotspot for research in the field of data modeling. Among the wide range of graph-based applications, graph similarity computation between pairs of graphs is a fundamental task, used in graph retrieval and clustering. Graph Edit Distance (GED) and Maximum Common Subgraphs (MCS) are the two commonly used domain-agnostic metrics to evaluate graph similarity in practice, by quantifying the minimum cost under optimal alignment between the entities of two graphs, such as nodes, edges, and subgraphs. Specifically, GED is defined as the minimum cost required to transform one graph into another, including inserting, deleting and substituting. MCS measures the similarity by identifying the largest common subgraph between the two graphs. In some cases, for example, when one graph is a subgraph of another, GED is equivalent to MCS \cite{bunke1997relation}. Unfortunately, computing exact GED or finding the largest subgraph in MCS is known to be a NP-hard problem. 

Classical GED methods such as $A^*$-Beam search (Beam) \cite{neuhaus2006fast}, Hungarian \cite{riesen2009approximate}, Bipartite \cite{fankhauser2011speeding}, etc, suffered from the time complexity. Recently, trainable neural network based models have been proposed to approximate the calculations of GED/MCS between the given pair of graphs. These neural network based methods embed the two graphs or sets of graph nodes into the same space and optimize the model parameters via supervision between the predictions and the ground truth, solving GED and MCS computations within a reasonable time. However, deep learning models are well-known for being recognized as 'black boxes', the typically characteristic ``one-to-one node/subgraph alignment`` in the classical computations of GED and MCS cannot be seen. Such alignment information provides interpretability for graph editing operations and visualizes the common subgraph between a pair of graphs, which is extremely useful for graph-based downstream tasks. Hence, researchers are encouraged to develop a matching graph similarity computation framework. However, it faced from the following challenges:

\begin{itemize}
\item \textbf{Lack of fine-grained labels.} More recently, some matching graph similarity computation models have applied explicit supervision to solve the optimal node level alignment problem \cite{piao2023computing, tan2021proxy, zhao2021ia, fey2020deep, roy2022interpretable}. However, labeling such fine-grained, domain-specific node-level is extremely labor-intensive, and existing graph similarity computation tasks primarily rely on the GED score as the similarity measure for query-corpus graph pairs. To be effective, an ideal matching graph similarity computation model needs to be carefully designed to satisfy the unsupervised fine-grained alignment labels.
\item \textbf{Need satisfying one-to-one node alignment.} The task of GED computation is to learn an optimal permutation matrix, in which nodes/edges/subgraph can be best aligned. GED computation is a quadratic assignment problem, more strictly, it is a bijection problem, meaning each entity in the graph has exactly one partner in the other graph and vice versa. Due to issues such as the lack of fine-grained labels, existing methods \cite{doan2021interpretable, roy2022interpretable, chen2020graph, petric2019got} ignore these strict one-to-one constraints, only approximate the alignment matrix to some extent, in this paper we refer to it as soft alignment. Soft alignment is not consistent with the classical GED computation and leads to a inferior result. To achieve hard alignment, it is necessary to satisfy the one-to-one requirement, which requires the permutation matrix to be both symmetric and doubly stochastic. These new limitations increase the complexity of the model design.

\item \textbf{Need differentiable.} The goal of graph matching neural network is to efficiently approximate the classical intractable methods, including similarity computation and permutation matrix (i.e., explanations) generation. However, compared to the soft alignment, the hard alignment function is implemented by obtaining the binary relevance score, creating many kinks which is non-differentiable.
\end{itemize}

In this paper, we propose a novel interpretable neural node alignment model (GNA) without relying on node alignment ground truth information to address the above challenges. We first relax the quadratic assignment problem in classical GED computation into a linear alignment in the learnable node embedding space via a graph neural network, such that the subgraph structure is encapsulated into these node's features. Secondly, GED is formulated as the minimum assignment cost required to transform one graph into another based on node alignment, which consists of a node-node cost matrix and a node-level alignment matrix. Each value in the cost matrix denotes the cost of inserting, deleting, and substituting nodes, while each value of $0/1$ in the alignment matrix indicates whether the node pair is matched. This paper generates this hard alignment matrix via the differentiable Gumbel-Sinkhorn module. Specifically, to solve the non-differentiable issues in hard alignment, we adopt the doubly stochastic Sinkhorn algorithm with the Gumbel function to facilitate a differentiable graph matching task. To the best of our knowledge, we are the first to attempt predicting the graph similarity score while providing one-to-one node matching results. Our method achieves superior performance in both similarity approximation and retrieval tasks. In summary, the main contributions of this paper are as follows:

\begin{enumerate}[(1)] 
\item This paper proposes a novel graph similarity computation framework, which treats graph similarity computation as a prediction task while simultaneously providing an explanation for alignment operations at the node level without fine-grained annotations. Compared with existing methods, our method is more similar to classical intractable methods.
\item This paper designs an efficient differentiable graph matching module that leverages the doubly stochastic Sinkhorn algorithm with the Gumbel function to address the non-differentiable issue in hard alignment. To the best of our knowledge, our model is the first that satisfies the one-to-one node matching requirements. 
\item A large number of experiments are carried out on three real world graph datasets to verify the effectiveness and efficiency of our method. The proposed algorithm achieve the superior performance in both similarity approximation and retrieval tasks. Furthermore, visualized experiments demonstrate the high interpretability of our proposed method.
\end{enumerate}

The rest of this paper is organized as follows. We present the most related work in Section \ref{sec-2}. Following, we describe the preliminaries of our method and the details of the GNA framework in Section \ref{sec-3} and \ref{sec-4}. Experimental results are provided in Section \ref{sec-5}. Finally, we conclude our discussion and future work in Section \ref{sec-6}.

\section{Related Work}\label{sec-2}
In this section, we introduce the related works about graph similarity computation and graph/node matching.
\subsection{Graph Similarity Computation}\label{sec-2.1}
\noindent\textbf{Traditional Search-based Similarity Computation.} These traditional methods rely on graph search and structural matching. Typically, the A* beam search method \cite{neuhaus2006fast} optimizes the search space via heuristic strategies that apply a cost function to estimate the minimal cost from the current node to the goal one. This strategies are beneficial to reduce the complexity of each search step by limiting the number of candidate pool into a controllable size. Other approaches such as Bipartite approximation \cite{fankhauser2011speeding, jonker1988shortest} also utilized greedy search strategies similar to A* beam, yet progressively approximating the potential optimal solution based on local optimal solutions. Additionally, subgraph isomorphism is also a hot direction in this domain, for example Lou et al. \cite{lou2020neural} proposed to identify the similar substructures between a given pair of graphs to explore the potential graph matching. These traditional search methods are high interpretability which is valuable for downstream tasks such as in bioinformatics and cheminformatics fields. Howvever, they also suffered from the exponential time complexity, limiting their applicability to large-scale graphs. 

\noindent\textbf{Neural Network-based Similarity Computation.} Considering that a desirable graph search method should achieve higher accuracy and improve interpretability within a reasonable time. Researchers are encouraged to design models based on neural network to solve the above problem. Early works about node-level embedding based methods such as Node2Vec \cite{grover2016node2vec} and DeepWalk \cite{perozzi2014deepwalk} encoded the node features based on random walk, then transforming the graph similarity computation into vector comparisons. Additionally, graph-level similarity computation have been proposed, for example SimGNN \cite{bai2019simgnn}, GraphSim \cite{bai2020learning}, Graph Matching Network (GMN) \cite {li2019graph}. Although these existing GNN-based methods achieve better generalization and significant time reduction in exact GED computation, they lack an explanation of node/subgraph alignment.

\subsection{Graph/Node Matching}\label{sec-2.2}
\noindent\textbf{Graph Matching.} The matching method is designed to enhance the interpretability of the above GNN-based models including (sub)graph-level and node-level matching. Problems such as entity matching \cite{zheng2024cross}, social network alignment \cite{peng2020motif}, and protein-protein interaction network \cite{zhao2023semignn} alignment can be classified as graph matching. SLOTAlign \cite{tang2023robust} proposed to learn the knowledge graph structure information to generate the the entities matching matrix. GOT \cite{chen2020graph} employed the Gromov-Wasstein distance to obtain the optimal permutation matrix. GNN-PE \cite{ye2024efficient} proposed a novel pruning strategy based on path labeling/significant features, which efficiently and accurately matches subgraphs by traversing indexes over graph partitions in parallel.

\noindent\textbf{Node Matching.} The goal of node-level matching is to find the correspondence between nodes, which can be classified into many-to-one (soft) and one-to-one (hard). Node matching is closely related to various graph tasks, including graph alignment \cite{yao2022entity}, GED computation \cite{yang2021noah}, and subgraph matching \cite{raveaux2010graph}. For example, GOTSim \cite{doan2021interpretable} propsed a linear matching algorithm to derive one-to-one node alignment from the node similarity matrix and the addition/deletion cost matrix generated by the graph neural network. Additionally, GEDGNN \cite{piao2023computing} designed the Cross Matrix Module to obtain soft matching, accompanied by fine-grained supervision for gradient learning.

\section{Preliminaries}\label{sec-3}
\noindent \textbf{Graph matching problem.} In the graph matching task, finding the optimal permutation matrix is known to be a NP-hard problem. Given two graphs adjacency matrix $A_q$ and $A_c$, in generally $G_q \subseteq G_c$, each non-zero element in $A_c$ can be found the only corresponding element in $A_q$ via the row and column permutation conversion. Take the case of $A_q$ and $A_c$ has same size as the example, we formula the subgraph isomorphism as the following optimizing problem:
\begin{align}\label{eq-1}
argmin_P {\textstyle \sum_{i,j}}\left [ \left ( A_q- PA_cP^{\top} \right )_+ \right ] _{i,j} 
\end{align}

Where $\left ( \bullet  \right ) _+$ denotes only values greater than $0$ will be reserved. The objective function illustrates how close $G_c$ to $G_q$ and the minimum objective is achieved under the optimal permutation matrix $P$. However, optimizing the formula (\ref{eq-1}) is NP-hard as it is a typical quadratic assignment problem, which is also one of the main challenges of GNA.

\noindent \textbf{Relax (1) to the linear assignment problem.} Formula (\ref{eq-1}) can be rewritten as the following equation (\ref{eq-2}), which suffers from the high time complexity. Let $H_q$, $H_c$ denote the node embedding of $G_q$ and $G_c$, respectively, and let $R_q$ and $R_c$ represent the node embedding of $G_q$ and $G_c$, respectively. The optimal assignment matrix $P$ needs to satisfy both the minimization of the node cost and edge cost, which is a notoriously difficult quadratic assignment problem.
\begin{align}\label{eq-2}
d(G_C|G_q)=argmin_P\left \{  {\textstyle \sum_{i,j}\left [ \left ( H_q-PH_c \right )_+  \right ]_{i,j} } +  {\textstyle \sum_{e}\left [ \left ( R_q-PR_c \right )_+  \right ]_{e} }  \right \}
\end{align}

However, we find that the individual components of formula (\ref{eq-2}) are linear assignment problems, which are easy to solve. To tackle the quadratic assignment problem, we utilize the node-to-node alignment (first part of the formula (\ref{eq-2})) to measure the relevance score between $G_c$ and $G_q$ in the node embedding space. This efficiently relaxes formula (\ref{eq-1}) to a linear assignment, reducing the assignment search to polynomial time. Thus, in this paper the objective of the subgraph isomorphism is updated as:
\begin{align}\label{eq-3}
d(G_C|G_q)=argmin_P  {\textstyle \sum_{i,j}\left [ \left ( H_q-PH_c \right )_+  \right ]_{i,j} } 
\end{align}

Although the latter part of formula (\ref{eq-2}) can still solve the quadratic assignment problem, in this paper, we utilize node features to generate the optimal graph alignment, as the node features generated by the GIN encoder encapsulate both the subgraph structure and edge feature information. Detailed information will be provided in Section \ref{sec-4.1}.

\section{Our Method}\label{sec-4}

\begin{figure}[h]
  \centering
    \includegraphics[width=0.9\textwidth]{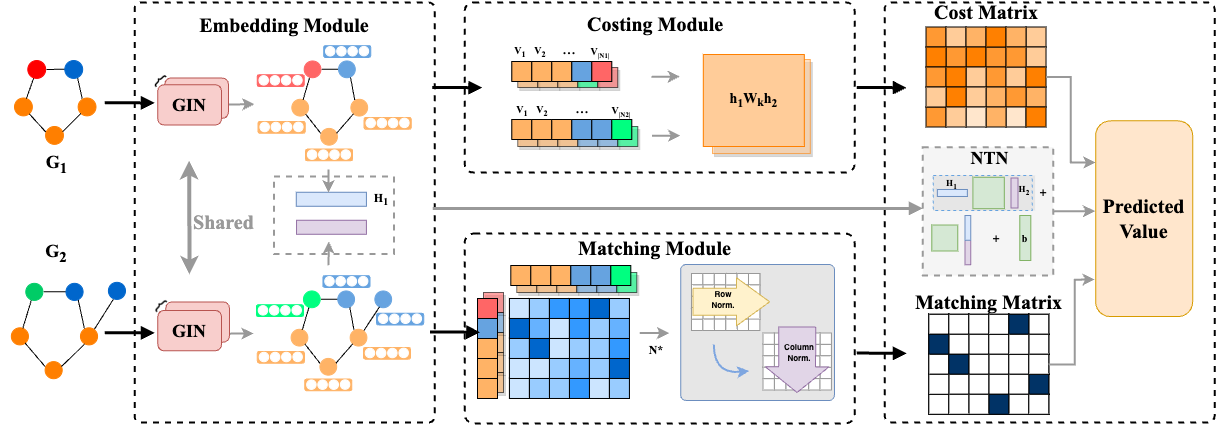}
    \caption{The framework of the GNA.}
    \label{fig1}
\end{figure}
The proposed interpretable neural node alignment model, GNA, which comprises three components: 1) Embedding Module and 2) Costing and Matching Module, 3) Prediction, as illustrated in Figure \ref{fig1}. For a given pair of graphs, GNA first utilizes the multiple GNN layers to encode the graph structure and feature information into the node embeddings. Then, two bilateral branches are executed in GNA. One branch aims to estimate the pairwise node-to-node matching cost named ``costing module''. The matching module outputs the optimal permutation matrix, as shown in Figure \ref{fig1}. When dealing with pairs of graphs that have different numbers of nodes, it is necessary to compute the associated costs of node addition or deletion under the one-to-one node matching setting. Thus, we pad the cost and matching matrices to make it square. Detailed information is provided in Section \ref{sec-4.2}. Finally, the outputs from the node-level costing and matching, along with the graph-level embeddings, are aggregated to predict the similarity score.

\noindent \textbf{Problem statement:} We focus on the undirected and unweighted graphs, which is denoted as $G=\{V,A\}$, where $v_i \in V$ is the node set and $A \in R^{N\times N}$ represents the graph adjacency matrix. Given a pair of graphs $G_1=\{V_1, A_1\}$ , $G_2=\{V_2, A_2\}$, generally $|V_1|\le|V_2|$, the graph similarity computation is defined to generate the similarity score $s(G_1, G_2) \in [0,1]$ based on the graph features. 
\subsection{Embedding Module}\label{sec-4.1}
Recently, graph neural networks (GNNs) have gained significant attention as a backbend for encoding node embeddings in various graph tasks. Following ref. \cite{piao2023computing}, we adopt three-layer of the Graph Isomorphism Network (GIN \cite{xu2018powerful}]) as the basic node encoder, which has been verified to have superior performance in different graph structural information capturing compared to conventional GNN models (e.g., GCN, GraphSAGE\cite{hamilton2017inductive}). 

Inspired by the Weisfeiler-Lehman isomorphism test, GIN achieves comparable results using a learnable multi-layer perception module. The $k-$th layer output of the GIN is formulated as follows:
\begin{align}\label{eq-4}
h_{v}^{(k)} = MLP^{(k)} ( ( 1+\varphi^{(k)} )h_{v}^{(k-1)} + {\textstyle \sum_{u \in N(v)} h_{v}^{(k-1)}} )
\end{align}

where $h_{v}^{(k)} \in R^d$ represents the representation of node $v$ at the k-th GIN layer, $N(v)$ denotes the set of neighbors of node $v$, $\varphi^{(k)}$ represents the trainable parameter. From equation \ref{eq-4}, it is easy to see that the aggregation function of GIN is sum aggregation, which propagates information from high-order neighbors and the node itself without normalization. $\varphi^{(k)}$ is designed to distinguish the similarity between the center node and different graph structures, which efficiently avoids over-smoothing.

In this work, for each node, the initial feature $h_{v}^{(0)} \in R^M$ is determined according to its labels. If the node has one label, $h_{v}^{(0)}$ will be a one-hot vector, more labels the corresponding dimensions will be set $'1'$. In the case of nodes without labels, we simply set $h_{v}^{(0)}$ as a random constant number. We then propagate the structural information using multiple layers of GIN and aggregate it using equation \ref{eq-4}.
\subsection{Costing Module}\label{sec-4.2}
As shown in Figure \ref{fig1}, the graph similarity computation is composed of two components. We elaborate on the costing and matching components at the node-level. Inspired by the bilateral branch network \cite{bai2019simgnn}, we adopt two separate branches to equip the model with both alignment and retrieval capabilities. From the implementation perspective, the dual branch is more consistent with the subsequent calculations, which will be explained in the following sections.

After the GIN backend, node-level embeddings of two graphs $H_1$ and $H_2$ are generated. The cost computation module is designed to estimate the transfer cost (e.g institution, add and delete) from one node to another between the graph pairs $(G_1, G_2)$. The goal of cost module is $\mathbf{A}_{ij}^{cost} \in [0,1]$ of size $|V_1|\times|V_2|$ in which an element $\mathbf{A}_{ij}$ denotes the cost of edit operations for matching $v_i \in V_1$ with $v_j \in V_2$. 
\begin{align}\label{eq-5}
\mathbf{A}_{ij}^{cost} & = 
\left[
\begin{array}{ccc} 
\mathbf{c}_{1,1} & \cdots & \mathbf{c}_{1, |V_{2}|}  \\
\vdots & \ddots & \vdots  \\
\mathbf{c}_{|V_{1}|, 1} & \cdots & \mathbf{c}_{|V_{1}|, |V_{2}|}
\end{array}
\right]
\end{align}

However, based on the formula (\ref{eq-5}), $\mathbf{A}_{ij}^{cost}$ is a symmetric matrix that only record the substitution cost of inter-node without the costs of node deletion and addition.  This motivates us to extend the matrix $H_1 \in R^{|V_1|\times d}$ as $H_1^* \in R^{|V_2|\times d}$. Several strategies have been proposed to solve this problem. A commonly used approach is padding a zero vector into $H_1$ \cite{roy2022interpretable}. In this study, $|V_2|-|V_1|$ nodes will be added to the smaller graph to ensure equal node sizes between the two graphs. The graph-level embedding will be used as the initial features for the newly added nodes. The graph-level embedding, generated through mean pooling of the node-level features, will serve as the initial features for these newly added nodes. Such a graph-level embedding enables the model to incorporate the graph structure into the node deletion cost computation. Since deletion cost is typically equivalent to the addition cost, we define the addition cost as the cost associated with editing unaligned nodes in this study. The revised cost matrix is formulated as follows, the rows below the horizontal line denotes the $|V_2|-|V_1|$ new added nodes in $G_1$.
 
\begin{align}\label{eq-6}
\mathbf A_{cost} & = 
\left[\begin{array}{ccc}
\mathbf{c}_{1,1} & \cdots & \mathbf{c}_{ 1, |V_{2}|}  \\
\vdots & \ddots & \vdots  \\
\mathbf{c}_{|V_{1}|, 1} & \cdots & \mathbf{c}_{|V_{1}|, |V_{2}|}\\
\hline 
\mathbf{c}_{|V_{1}|+1,1} & \cdots & \mathbf{c}_{|V_{1}|+1, |V_{2}|}\\
\cdots & \cdots & \cdots \\
\mathbf{c}_{|V_{2}|,1} & \cdots & \mathbf{c}_{|V_{2}|, |V_{2}|}
\end{array}\right]
\end{align}

Multiple linear functions without bias are adopted to empower the expressivity of the model. $H_1^*$ represents the node embeddings after padding. Unless the inner product used in formula (\ref{eq-7}), other measures such as Euclidean distance are also be used to generate the cost matrix.
\begin{align}\label{eq-7}
\mathbf A = H_1^* W H_2^\top
\end{align}

Multiple layers of Equation (7) are stacked to compute the final matching cost between the nodes of the two graphs through iterative interactions, which has been proven to be beneficial to the model's expressive ability in the work \cite{doan2021interpretable}.
\begin{align}\label{eq-8}
\mathbf A_{ij}^{cost} = \left [ H_1^* W_1 H_2^\top, H_1^* W_2 H_2^\top,..., H_1^* W_l H_2^\top \right ] 
\end{align}

\subsection{Matching Module}\label{sec-4.3}
Matching module outputs optimal permutation matrix $A_{ij}^{matching} \in \{0,1\}^{|V_2|\times|V_2|}$ where each element $A_{ij}^{matching}$ indicates whether the node $v_{i}$ in $G_1$ are matched to the node $v_{j}$ in $G_2$, and the $|V_2|-|V_1|$ nodes in $G_2$ will be deleted (or added). The optimal permutation matrix $A^*$ is a doubly stochastic matrix that satisfies the one-to-one matching in the linear assignment task. As the objective function in the formula (\ref{eq-3}), we apply the $Relu$ function to ensure $(\bullet)_+ $. Specifically, a $Linear-ReLU-Linear$ network named ``LRL'' is performed on $H_1^*$ and $H_2$. LRL is utilized to generate the costing matrix as the foundation for determining pair of node matches, which is similar to the costing module.
\begin{align}\label{eq-9}
\mathbf A = GS(LRL_{\theta}(H_1) \cdot LRL_{\theta}(H_2))
\end{align}

From the GED definition, the optimal assignment matrix is a doubly stochastic. However, $A_{ij}^{matching}=0$ will make the neural cells non-differentiable. In this work, we propose to incorporate the Gumbel function with the Sinkhorn operator to find the optimal $A$. As in the Figure \ref{fig1} shown, the permutation matrix $A$ is approximated via multiple alternating iterations over the rows and columns as following:
\begin{align}
GS(M): & = \lim_{t \to \infty} GS^t (M), \quad where \label{eq-10} \\
GS^0(M): & = exp(M/\tau )  \quad and \label{eq-11} \\
GS^t(M): & = ColSacle(RowSacle(GS^{t-1}(M))) \label{eq-12}
\end{align}

Specifically, the GS network progressively transforms the input matrix $M$ into a doubly stochastic matrix by alternately performing row and column normalization. This process also regulates the smoothness and discreteness of the resulting approximate permutation matrix by adjusting the temperature parameter $\tau > 0$.

\subsection{Prediction}\label{sec-4.4}
$A_{ij}^{matching}$ is the final permutation matrix that demonstrates the value of the node matching based on the semantic distance between two nodes, $A_{ij}^{cost}$ is the edit cost matrix. Following existing methods \cite{piao2023computing}, the element-wise multiplication and summation is employed as the prediction GED of transforming one graph into another.Since both the permutation matrix and the cost matrix are obtained via the inner product based on the node embedding, an NTN network is designed to calculate a bias in the prediction based on the graph-level features which serve as supplement information. This strategy is widely used in several GNN-based networks \cite{bai2019simgnn, zhuo2022efficient, piao2023computing} that rely on node-to-node cost to predict the graph similarity. 
\begin{align}\label{eq-13}
GED(G_1,G_2)_{prediction}=A_{ij}^{matching}*A_{ij}^{cost}+bias  
\end{align}
Due to similarity score ranging in $[0,1]$, we normalized the prediction GED by the sigmoid function as follows:
\begin{align}\label{eq-14}
Pred\_ Score(G_1,G_2)=\sigma (A_{ij}^{matching}*A_{ij}^{cost}+bias)
\end{align}
Similarly, we also normalized the ground-truth GED into $[0,1]$. Several normalized methods have been proposed in previous studies. In this paper, we utilize the linear function to generate the ground truth of the similarity as follows:
\begin{align}\label{eq-15}
GT\_Score(G_1, G_2)=\frac{GED(G_1, G_2)}{max(|V_1|,|V_2|)+max(|E_1|,|E_2|)}
\end{align}
From the formula (\ref{eq-15}), the denominator is the original GED value, the similarity score will be located in $[0, 1]$. After this conversion, we can simply use the minimum squared error (MSE) to compute the training loss:
\begin{align}\label{eq-16}
\mathcal L = \frac{1}{B} \sum_{(G_1,G_2) \in B} ||Pred\_ Score(G_1,G_2)-GT\_Score(G_1, G_2)||_2^2
\end{align}

\begin{algorithm}
\caption{GNA Model Training}
\label{alg1}
    \begin{algorithmic}[1]
    \State \textbf{Input:} Training graph pairs $(G_1,G_2)$, number of training epochs $K$, learning rate $\eta$, size of batch $m$ 
    \State \textbf{Output:} parameters of the network $\{\Theta\}$
    \For {number of training epochs K }
        \State Sample a mini-batch $\mathit{\{(G_1, G_2)_u | u= 1, ... , m\}}$ of graph pairs.
        \For{ each graph pair $(G_1, G_2)$ }
            \State Compute node embeddings $h_{G_{1},i}$ and $h_{G_{2},j}$ $\forall v_i \in G_1, v_j \in G_2$.
            \State Compute cost matrix $A_{cost}$ using Eq.(8).
            \State Compute matching matrix $A_{matching}$ using Eq.(9).
            \State Compute $s(G_1,G_2)$ in Eq.(14) using $A_{cost}$ and $A_{align}$.
        \EndFor
        \State Compute the mini-batch loss $\mathcal L$ using Eq. (16).
        \State Update $\theta \gets \theta - \eta \frac{\partial \mathcal L}{\partial \theta} $
    \EndFor
    \end{algorithmic}
\end{algorithm}

\section{Experiments}\label{sec-5}
\subsection{Datasets}\label{sec-5.1}
We introduce three real-world graph datasets: \textbf{AIDS}, \textbf{Linux} and \textbf{IMDB} to evaluate our method, the detail descriptions are listed in the table \ref{tab1}. Here, '\#' denotes the total number, such as \#Graphs represents the number of graphs contained in the current dataset. We construct the pair graph training dataset following {\cite{ xu2021graph, bai2019simgnn}} as denoted as \#Pairs. 

Each graph in \textbf{AIDS} dataset is antivirus screen chemical compounds, consisting of 42,687 chemical compound structures with $C$, $N$, $O$, $Cu$, etc. atoms and covalent bonds. We selected graphs with the number of node and edges ranging from 2 to 10 and 1 to 14, respectively. Then randomly grouped them to generate the training set of 940,000 graph pairs.

\textbf{Linux} is a collection of 48,747 Program Dependence Graphs (PDGs) obtained from the Linux kernel functions. Each node in a graph represents a function statement, and each edge represents the dependency between two statements. We filter out graphs with more than 10 nodes and randomly select 1,000 graphs as the initial dataset.

\textbf{IMDB} is a larger dataset in which contains a maximum of 89 nodes. The \textbf{IMDB} dataset consists of a ego-network of 1,500 movie actors, with each node representing an actor and each edge a co-star relation. To avoid significant differences in the size of the graph, for \textbf{IMDB} large graph (number of nodes > 10), we generate 100 synthetic graphs following the ref. \cite{piao2023computing}. For smaller graph, pairwise combinations is adopted. Finally, there are 501 × 501 + 399 × 100 graph pairs generated for the model training in total.

We split the three entire dataset with a ratio of 6:2:2 into training, validation, and testing sets.

\subsection{Experiment Setting and Metrics}\label{sec-5.2}
\subsubsection{Hyper-parameter Settings}\label{sec-5.2.1}
We stack $3-$layer GIN as basic graph encoder following the GEDGNN\cite{piao2023computing} and the output feature size of each layer is 32, 64 and 128 respectively. For the cross matrix module, we set the number of layers as 16. For the matching matrix module, the temperature parameter is 0.1. The training batchsize is set to 128, all methods achieved the best result with less 20 epochs. We set the learning rate to 0.001 with the Adam optimizer, and the weight decay is 5 × $10^{-4}$.

\begin{table}[htbp]
    \caption{Statistics of Graph Data Sets} 
    \begin{tabular}{c | c c c c c c} 
        \toprule
          & $\#$Graphs & $\#$Pairs & Avg$\#$Nodes & Avg$\#$Edges & $\#$Nodes & $\#$Edges \\
        \midrule
        \textbf{AIDS}       & 700   & 490K      & 8.9   & 8.80  & \makecell{$2\le N \le 10$ } & \makecell{$1\le N \le 14$ } \\
        \textbf{Linux}       & 1000  & 1M        & 7.6   & 6.94  & \makecell{$4\le N \le 10$ } & \makecell{$3\le N \le 13$ } \\
        \textbf{IMDB}        & 1500  & 289.9M    & 19.1  & 117.1 & \makecell{$7\le N \le 89$ } & \makecell{$12\le N \le 1467$ } \\
        \bottomrule
    \end{tabular}
    \label{tab1}
\end{table}

All experiments are done on a Windows PC using the i7-8750H CPU. Our model is implemented in Python 3.7, with machine learning models developed using PyTorch and PyTorch Geometric. For a fair comparison, all baselines are performed using the reported hyperparameter settings.
\subsubsection{Baselines}\label{sec-5.2.2}
We evaluate the proposed method against 7 baselines, including end-to-end deep learning based approaches and node matching approaches.

\textbf{1. End-to-end GED prediction approaches.} (a) The EmbMean and EmbMax methods are the basic trainable models based on graph neural network by mean or max the node-level embeddings to predict the GED. (b) SimGNN\cite{bai2019simgnn} is the first and representative graph similarity method. (c) TaGSim \cite{bai2021tagsim} predicts the GED at a fine-grained level by defining the different graph edit types and convert the GED into the sum cost of each editing type. (d) GPN \cite{yang2021noah} is one of the state-of-art method in terms of graph edit path searching and graph similarity prediction, which is originally designed to supervise the A*-beam search. 

\textbf{2. Node matching algorithms.} (a) GOTSim\cite{doan2021interpretable} proposed an efficient unsupervised differentiable GED computation method and simultaneously learning the alignments (i.e., explanations) similar to those of classical intractable methods. (b) GEDGNN\cite{piao2023computing} is the the state-of-art method in node matching methods, however it is a supervised matching algorithm.

\subsubsection{Metrics}
We adopt the following commonly used metrics to valuate the performance between our model and baselines.

\textbf{1. Metrics about graph similarity estimation.} The goal of the graph similarity estimation is to evaluate the precision of the predicted GED. We consider the Mean Absolute Error (MAE) and Accuracy as the base metrics. For example, given a pair of testing graph, MAE is defined to calculate the distance between the predicted GED value $d$ and the ground-truth $d^*$ as $|d-d^*|$. Accuracy is a commonly used in the classification tasks, in this work, we round the predicted floating GED to a integer as $round(d)=d^*$ to evaluate the performance.

\textbf{2. Metrics about graph retrieval ranking.} Ranking Metrics are sensitive the item rank. There are 100 candidate graphs randomly selected from the item pool for each query graph $G$. Then, rank these candidates based on their predicted GED values. Therefore, we adopt Spearman’s Rank Correlation Coefficient denoted as $\rho$, Kendall’s Rank Correlation Coefficient denoted as $\tau$, and precision $(P@10, P@20)$ as the retrieval ranking evaluation metrics. 
\subsection{Performance Analysis}\label{sec-5.3}
\begin{table}[h]
    \centering
    \setlength{\tabcolsep}{5pt} 
    \renewcommand{\arraystretch}{1.2} 
    \caption{Presents the performance of all methods on the three real-world graph dataset. The best results are highlighted in bold.}
    \raggedright 
    \scalebox{0.8}{ 
        \begin{tabularx}{\textwidth}{@{} c|c c|c c c|c c|c c c|c c|c c c @{}}
            \cline{1-16}
            \multicolumn{1}{c|}{} & \multicolumn{5}{c|}{\textbf{AIDS}} & \multicolumn{5}{c|}{\textbf{Linux}} & \multicolumn{5}{c}{\textbf{IMDB}}\\
            \cline{2-16}
            \multicolumn{1}{c|}{\textbf{Models}} & MAE & Accuracy & $\rho$ & $\tau$ & P@10 & MAE & Accuracy & $\rho$ & $\tau$ & P@10 & MAE & Accuracy & $\rho$ & $\tau$ & P@10\\
            \cline{1-16}
            EmbMean & 1.024 & 0.321 & 0.773 & 0.631 & 0.573 & 1.069 & 0.346 & 0.767 & 0.670 & 0.798 & 8.421 & 0.087 & 0.597 & 0.512 & 0.496 \\
            EmbMax & 1.716 & 0.183 & 0.684 & 0.544 & 0.502 & 4.207 & 0.060 & 0.369 & 0.313 & 0.317 & 8.769 & 0.078 & 0.558 & 0.488 & 0.621 \\
            SimGNN & 0.914 & 0.338 & 0.832 & 0.693 & 0.624 & 0.456 & 0.596 & 0.933 & 0.844 & 0.891 & 1.470 & 0.232 & 0.480 & 0.364 & 0.578 \\
            TaGSim & 0.841 & 0.366 & 0.850 & 0.715 & 0.646 & 0.391 & 0.668 & 0.924 & 0.837 & 0.816 & 3.752 & 0.074 & 0.115 & 0.090 & 0.433 \\
            GPN    & 0.902 & 0.353 & 0.822 & 0.684 & 0.586 & 0.135 & 0.884 & 0.962 & 0.898 & 0.956 & 1.591 & \underline{0.277} & 0.551 & 0.456 & 0.546 \\
            \cline{1-16}
            GOTSim & 0.794 & 0.387 & 0.862 & 0.729 & 0.671 & 0.592 & 0.501 & 0.928 & 0.840 & 0.890 & 5.364 & 0.165 & \underline{0.695} & \underline{0.592} & \underline{0.752} \\
            GEDGNN & \underline{0.773} & \underline{0.397} & \underline{0.876} & \underline{0.751} & \underline{0.716} & \underline{0.094} & \underline{0.917} & \underline{0.963} & \textbf{0.903} & \underline{0.962} & \textbf{1.398} & 0.269 & 0.629 & 0.500 & 0.696 \\
            \cline{1-16}
            GNA & \textbf{0.609} & \textbf{0.459} & \textbf{0.901} & \textbf{0.781} & \textbf{0.759} & \textbf{0.056} & \textbf{0.948} & \textbf{0.964} & \underline{0.884} & \textbf{0.980} & \underline{1.425} & \textbf{0.406} & \textbf{0.759} & \textbf{0.631} & \textbf{0.846} \\ 
            Improve(\%) & 21.22 & 15.62 & 2.85 & 3.99 & 6.01 & 40.43 & 3.38 & 0.10 & -2.10 & 1.87 & -1.93 & 46.57 & 9.21 & 6.59 & 12.50 \\
            \cline{1-16}
        \end{tabularx}
    }\label{tab2}
\end{table}
In this section, we present the results of graph similarity estimation and graph retrieval ranking. Table \ref{tab2} show the similarity score estimation and ranking results for the GED problem, respectively. Our method outperforms most of the benchmarks on all metrics across the three dataset both in GED prediction task and graph retrieval task. GPN, GOTSim, GEDGNN achieved the second best performance in some dataset, respectively, which denotes our model can learn a good embedding function that generalizes to unseen test graphs. In \textbf{Linux} and \textbf{IMDB} dataset, GEDGNN outperforms in the terms of $\tau$ and $P@10$. The reason might be that GEDGNN utilize the node matching ground-truth targets, which is computationally very expensive. This can also be obtained from GOTSim, unless the graph matching task in the \textbf{IMDB} dataset, GEDGNN achieved better performance. It is worth mentioning that interpretable graph similarity computation models generally outperform traditional graph similarity computation methods (e.g., SimGNN), which indicates that interpretable models, either node matching or subgraph matching, can effectively improve model's understand ability about the differences between two pair graphs and improve the accuracy of GED. Although the embmean and embmax method achieves the worst results in Table \ref{tab2}, these basic methods provide the direction for the subsequent series of model studies.

\subsection{Ablation Study}\label{sec-5.4}
\begin{figure}[h]
    \centering
    \begin{subfigure}[t]{0.23\textwidth}
        \centering
        \includegraphics[width=\textwidth]{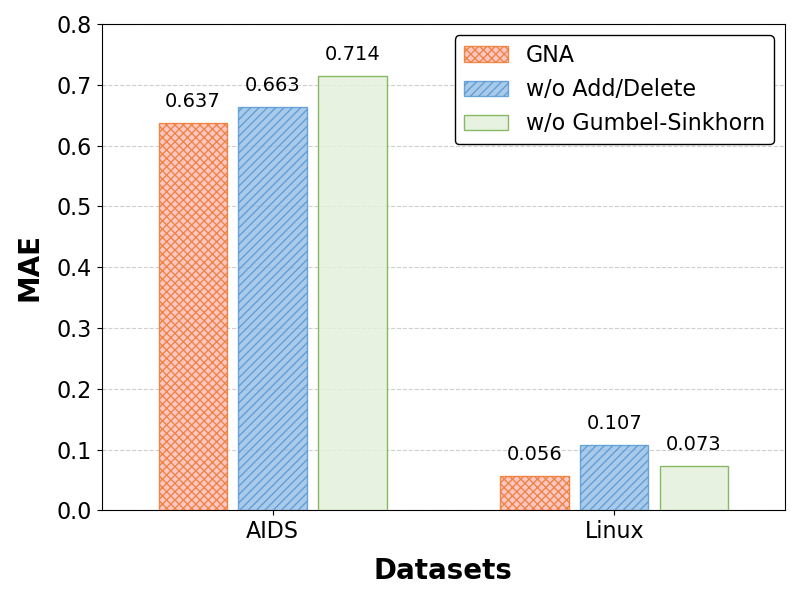}
        \caption{}
    \end{subfigure}
    \begin{subfigure}[t]{0.23\textwidth}
        \centering
        \includegraphics[width=\textwidth]{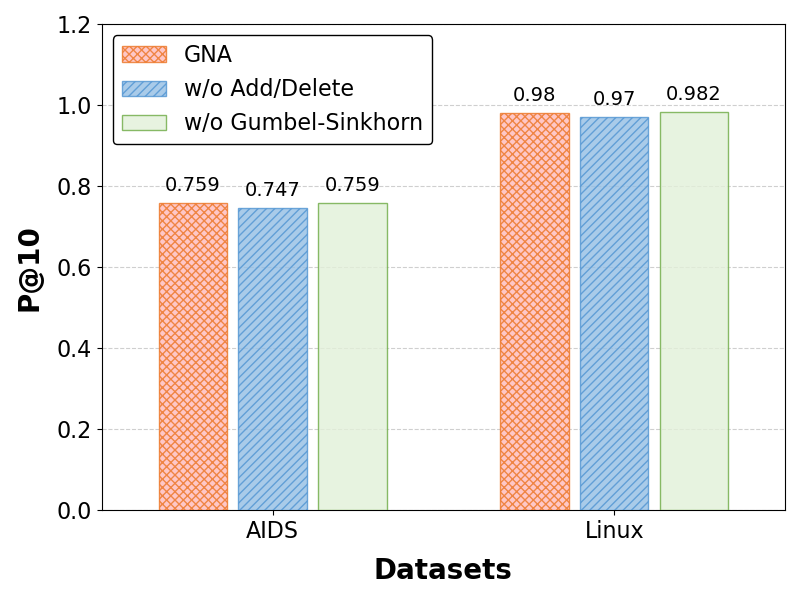}
        \caption{}
    \end{subfigure}
    \begin{subfigure}[t]{0.23\textwidth}
        \centering
        \includegraphics[width=\textwidth]{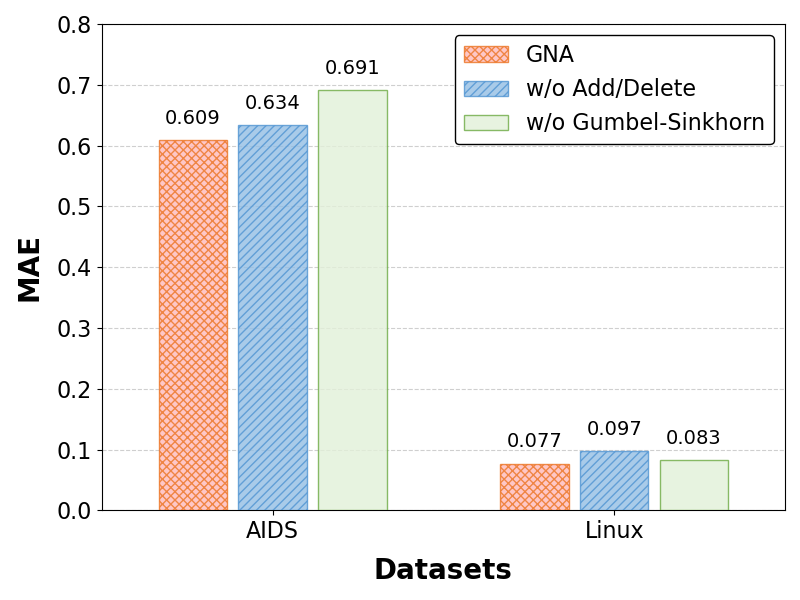}
        \caption{}
    \end{subfigure}
    \begin{subfigure}[t]{0.23\textwidth}
        \centering
        \includegraphics[width=\textwidth]{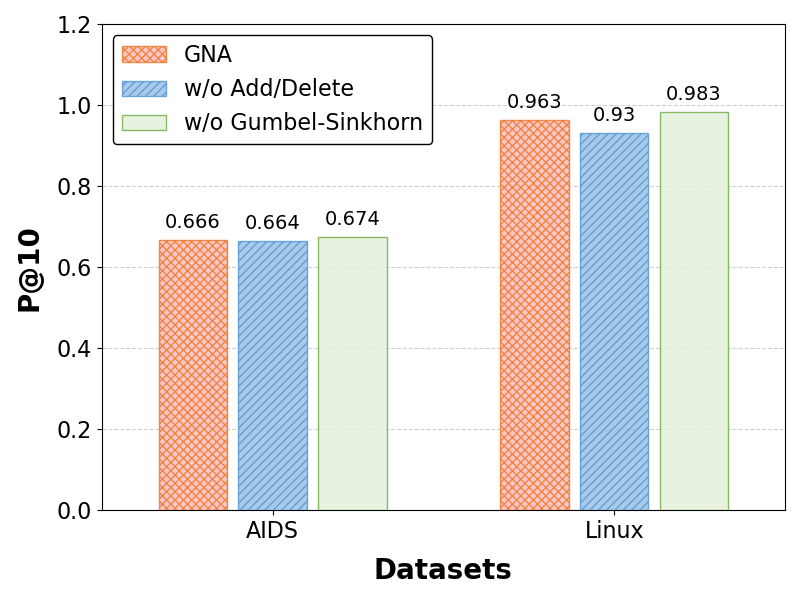}
        \caption{}
    \end{subfigure}
    \caption{Ablation study result. The top subfigures denotes on the original test set, the bottom ones are conducted on the filtered dataset.}
    \label{fig2}
\end{figure}

In this section, we evaluate the affect of the two key components in GNA: bijection in the matching module and delete/add cost in the cost module. w/o Add/Delete Cost refers to removing the add and deleting cost; w/o Gumbel-sinkhorn refers to removing the gumbel-sinkhorn algorithm to relax the bijection limitation. The final result is shown in Figure \ref{fig2}. In other words, the alignment matrix is obtained based on the node distance. To gain a deeper understanding of the affectiveness, we conducted two ablation studies, one is on the default testing dataset and the other testing dataset we filtered out pair graphs that $|V_2|=|V_1|$. In Figure \ref{fig2}, we remove the Gumbel-Sinkhorn algorithm to relax the bijection limitation. From the Figure \ref{fig2}, it is observed that the results demonstrate a significant decline in performance on the graph similarity similarity task, while showing slight variations on graph retrieval ranking metrics. It might be concluded that the similarity similarity task is more sensitive to the node-level alignment, while the graph retrieval task is vice versa. In spite of we generate two separate testing set on the similarity task, the overall trend is consistent that is GNA performs the best, w/o Add/Delete Cost following, w/o Gumbel-Sinkhorn performs the worse. It demonstrates that node-level interpretable especially bijection can effectively improve model's understand ability about the differences between two pair graphs and improve the accuracy of GED.

\subsection{Case study about Graph Ranking}\label{sec-5.5}
\begin{figure}[h]
    \centering
    \begin{subfigure}[t]{0.95\textwidth}
        \centering
        \includegraphics[width=\textwidth]{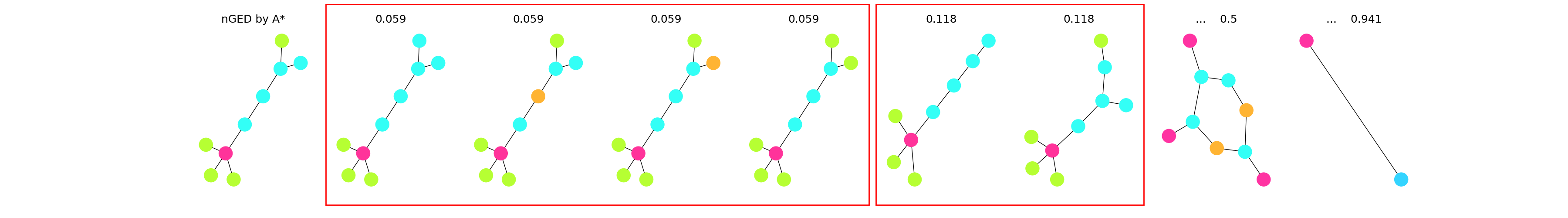}
    \end{subfigure}
    \vspace{1em} 

    \begin{subfigure}[t]{0.95\textwidth}
        \centering
        \includegraphics[width=\textwidth]{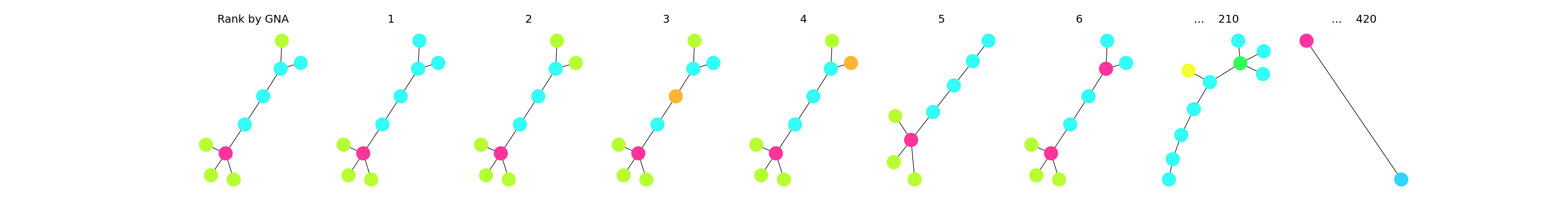}
    \end{subfigure}
    \vspace{1em}

    \begin{subfigure}[t]{0.95\textwidth}
        \centering
        \includegraphics[width=\textwidth]{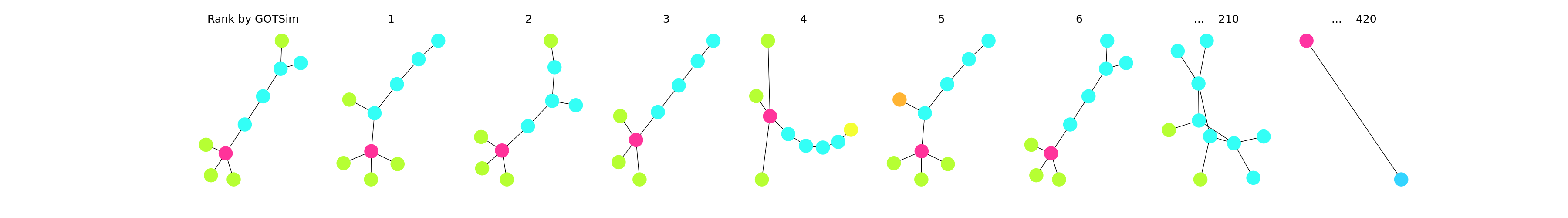}
    \end{subfigure}
    \vspace{1em}

    \begin{subfigure}[t]{0.95\textwidth}
        \centering
        \includegraphics[width=\textwidth]{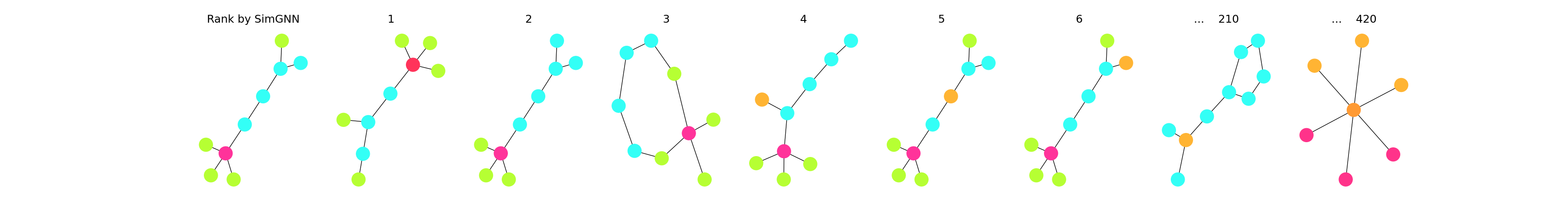}
    \end{subfigure}

    \caption{Visualizations of graph ranking result on \textbf{AIDS}. Nodes with the same color have the same label.}
    \label{fig3}
\end{figure}

We present the ranking results on the \textbf{AIDS} datasets with $A^*$, SimGNN (representative GNN based methods), GoTSim (unsupervised node-level matching method) and GNA. The top row represents the ground-truth ranking results by $A^*$ along with the predicted normalized GEDs. The graphs within the red rectangle are isomorphic, indicating that they have the same GED relative to the query graph. From Figure \ref{fig3}, it is clear that retrieve graphs obtained by GNA approximates to the ground truths. Furthermore, we obtain the same conclusions as shown in Table \ref{tab2}, that is integrating node matching exploring in the graph similarity computation task not only enhances the interpretability of the model but also improves its accuracy.

\subsection{Interpretable Graph Matching}\label{sec-5.6}
\begin{figure}[h]
    \centering
    \begin{subfigure}[t]{0.18\textwidth}
        \centering
        \includegraphics[width=\textwidth]{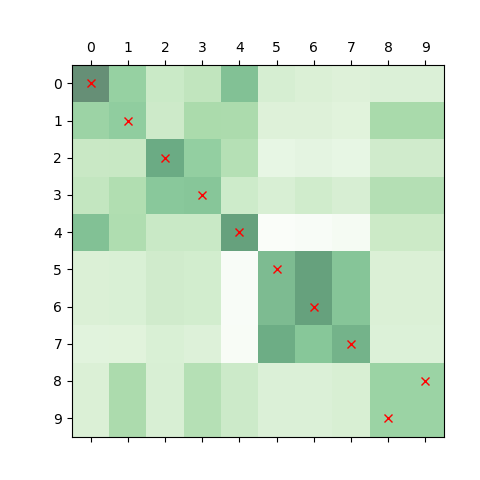}
    \end{subfigure}
    \begin{subfigure}[t]{0.28\textwidth}
        \centering
        \includegraphics[width=\textwidth]{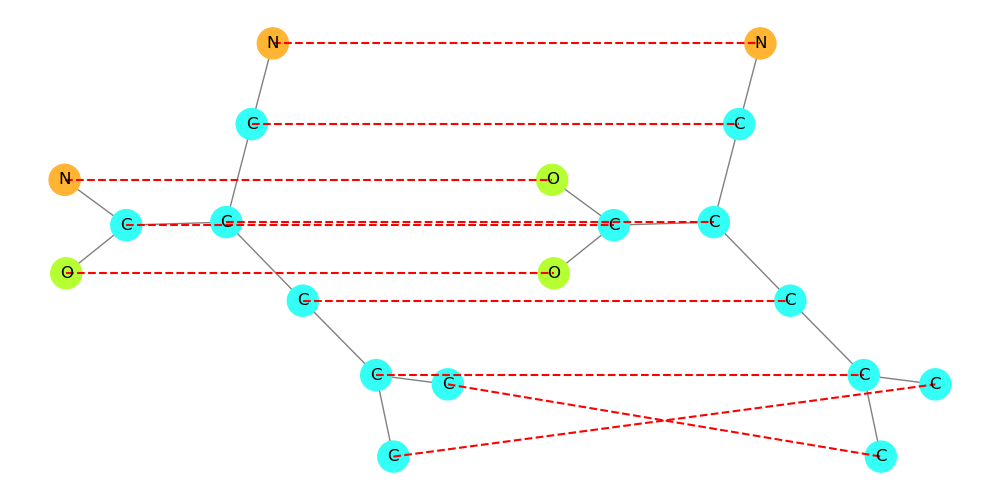}
    \end{subfigure}
    \begin{subfigure}[t]{0.18\textwidth}
        \centering
        \includegraphics[width=\textwidth]{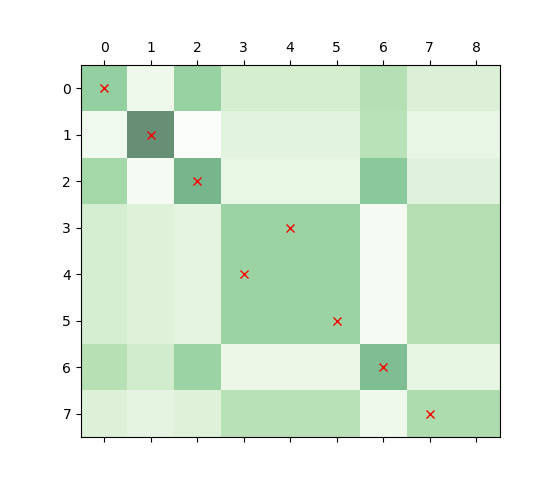}
    \end{subfigure}
    \begin{subfigure}[t]{0.28\textwidth}
        \centering
        \includegraphics[width=\textwidth]{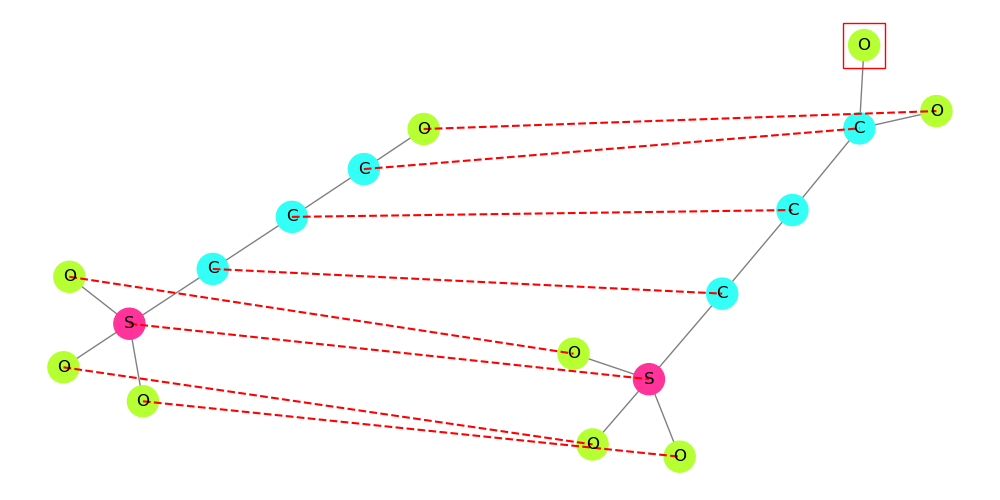}
    \end{subfigure}
    \caption{A graph matching case study for GNA on \textbf{AIDS} datasets. Heatmap denotes the final matching matrix at the node-level based on the node features generated by matching module. The depth of the color in the heat map indicates the weight of the node matching. The deeper the color, the more matched it is. The red rectangular boxes in the right of the visualize result denotes the node needed to be deleted.}
    \label{fig4}
\end{figure}

In this section, we discuss the effectiveness of GNA in the interpretable of the GED estimation and the results is shown in Figure \ref{fig4}. We conducted two case studies where one pair of graphs had the same number of nodes (the top row in Figure \ref{fig4}) and the other pair did not (the bottom in Figure \ref{fig4}). The matching matrix $A_{match}$ produced by our method is visualized as a heat map. The optimal node alignment is identified by ``$*$`` in Figure \ref{fig4}. GNA approximates the ground-truth node matching between a pair of graphs. Different from previous soft graph matching algorithms such as GMN \cite{li2019graph}, we can easily find the required operations to transform one graph into another based on the hard bijective mapping. This implies that GNA is highly interpretable and more similar to the classical GED models.

\section{Conclusion}\label{sec-6}
The paper proposes a novel interpretable graph similarity computation framework that provides node-level bijection alignment in GED approximation tasks. First, this architecture employs multi-layers of GIN to encapsulate the subgraph and edge information into the node features. Thanks to the embedding layer, the classical quadratic assignment problem is relaxed to a linear alignment. Then costing and matching modules are adopted to enhance the interpretable ability of the GED prediction. Experimental results verify that fine-grained alignment is beneficial to the expressive ability of the model. GNA achieves superior performance both in the GED prediction task and graph retrieval task. In this work, GNA embeds the node features to relax the quadratic assignment problem. Additionally, edge embedding presents an alternative solution. In the future, we will attempt to achieve the edge alignment in the graph similarity computation. More attention will be paid to find the solution about the graph similarity and entity alignment on more complex graphs, such as knowledge graphs and heterogeneous graphs.




\printcredits

\bibliographystyle{elsarticle-num}

\bibliography{ref}



\end{document}